\documentclass[10pt,twocolumn,letterpaper]{article}

\usepackage{cvpr}              

\usepackage{makecell}

\definecolor{cvprblue}{rgb}{0.21,0.49,0.74}
\usepackage[pagebackref,breaklinks,colorlinks,allcolors=cvprblue]{hyperref}

\def\paperID{8858} 
\def\confName{CVPR}
\def\confYear{2025}

\title{RSUniVLM: A Unified Vision Language Model for Remote Sensing via Granularity-oriented Mixture of Experts}

\author{Xu Liu, Zhouhui Lian\\
Wangxuan Institute of Computer Technology\\
Peking University, Beijing, China\\
{\tt\small  \{lxmyr, lianzhouhui\}@pku.edu.cn}
}

\makeatletter
\let\@oldmaketitle\@maketitle%
\renewcommand{\@maketitle}{\@oldmaketitle%
 \centering
    \includegraphics[width=\textwidth]{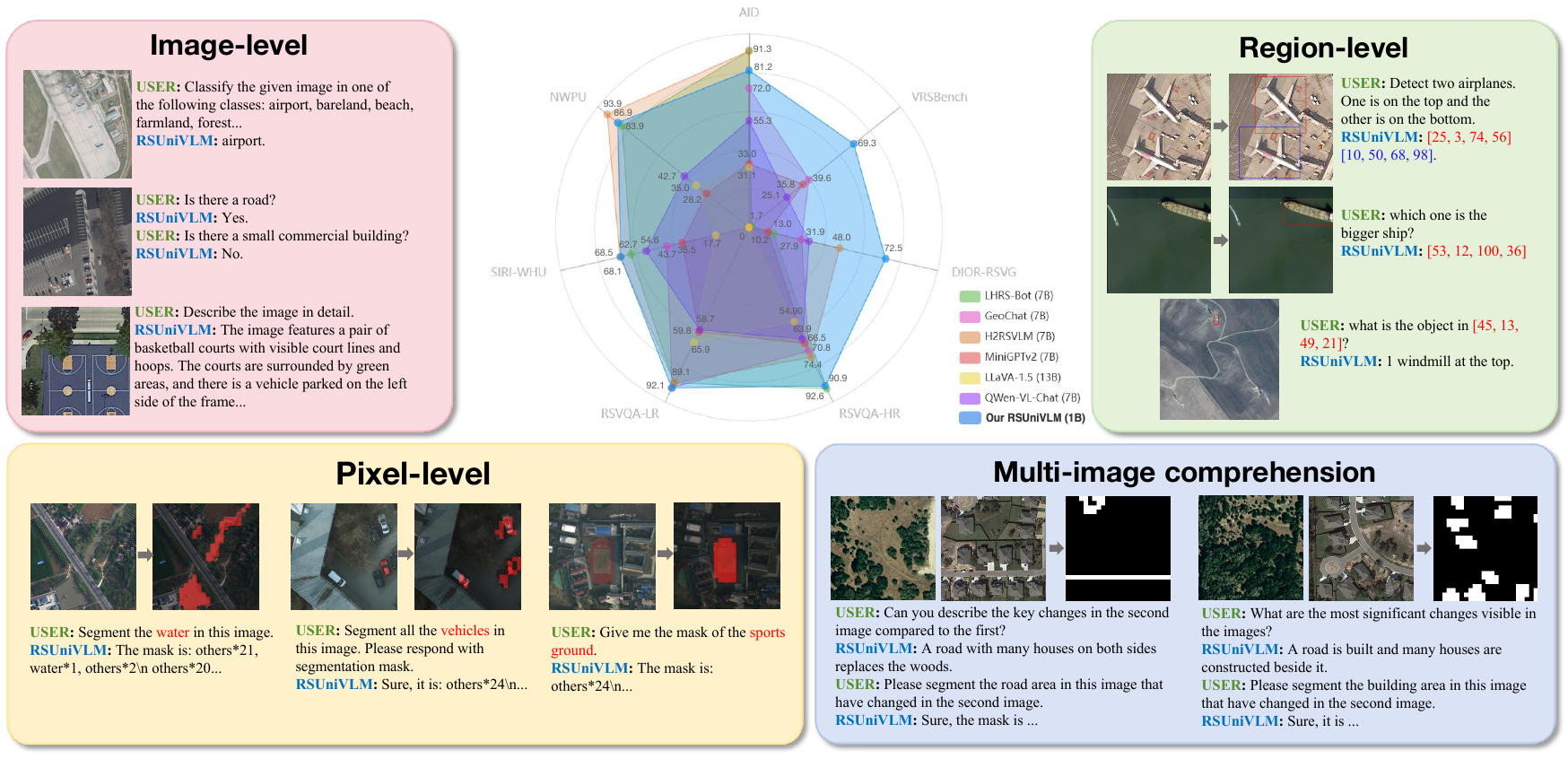} 
     \captionof{figure}{RSUniVLM is a unified remote sensing VLM with versatile capabilities across three levels of visual understanding: a) Image captioning and visual question answering at image-level; b) Visual grounding and referring expression generation at region-level; c) Semantic segmentation at pixel-level. Apart from tasks with single-image inputs, our RSUniVLM can also tackle multi-image comprehension tasks, such as change captioning and change detection. The radar chart demonstrates that our model is competitive with others on most datasets, and performs significantly better on VRSBench and DIOR-RSVG.}
    \label{fig:teaser}
    \bigskip}
\makeatother
\usepackage{multirow}
\usepackage{bbding}
\begin{document}
\maketitle

\begin{abstract}
Remote Sensing Vision-Language Models (RS VLMs) have made much progress in the tasks of remote sensing (RS) image comprehension. While performing well in multi-modal reasoning and multi-turn conversations, the existing models lack pixel-level understanding and struggle with multi-image inputs. In this work, we propose RSUniVLM, a unified, end-to-end RS VLM designed for comprehensive vision understanding across multiple granularity, including image-level, region-level, and pixel-level tasks. RSUniVLM also performs effectively in multi-image analysis, with instances of change detection and change captioning. To enhance the model's ability to capture visual information at different levels without increasing model size, we design a novel architecture called Granularity-oriented Mixture of Experts to constraint the model to about 1 billion parameters. We also construct a large-scale RS instruction-following dataset based on a variety of existing datasets in both RS and general domain, encompassing various tasks such as object localization, visual question answering, and semantic segmentation. Substantial experiments have been conducted to validate the superiority of the proposed RSUniVLM up to state-of-the-art across various RS tasks. Code and model will be available at \href{https://github.com/xuliu-cyber/RSUniVLM}{here}.
\end{abstract}    
\section{Introduction}
\label{sec:intro}

In the past few years, the emergence of Large Language Models (LLMs) \cite{radford2018improving,radford2019language,guu2020retrieval,raffel2020exploring} has driven remarkable development across numerous fields. By equipping with vision encoder module \cite{radford2021learning,zhai2023sigmoid}, Large Vision-Language Models (LVLMs) \cite{liu2024visual,zhu2023minigpt,alayrac2022flamingo,chen2024far,bai2023qwen} extend the capabilities of LMMs to general-purpose visual and language comprehension. A groundbreaking work is LLaVA \cite{liu2024visual}, demonstrating excellent visual chat capabilities when fine-tuned on multimodal chat data. To support a broad range of visual tasks, subsequent works \cite{bao2022vlmo,li2024unifiedmllm,ma2025groma} try to exploit potentialities of LVLMs in various ways, including leveraging instruction tuning data of larger scale and higher quality. \cite{chen2023sharegpt4v,xiao2024florence,xu2022multiinstruct}, designing more efficient fine-tuning methods \cite{hu2021lora,dettmers2024qlora,xu2023qa} and adopting novel LLM architectures \cite{lin2024moe,lin2023sphinx,zhang2023robust}. Moreover, some latest works \cite{zhang2024omg,wu2024visionllm,li2024unifiedmllm,fei2024vitron} explore the flexibility of LVLMs to unify the multi-modal perception and generation tasks, with the utilization of task-specific heads. 

Although the generic LVLMs have shown impressive capabilities in general domain, their performance in the remote sensing (RS) domain is underwhelming because of the significant gap between RS images and natural scene images. To bridge this gap, several large scale RS image-text pairs datasets \cite{zhang2024rs5m,ge2024rsteller,muhtar2024lhrs} and instruction tuning datasets \cite{kuckreja2024geochat,muhtar2024lhrs,pang2024h2rsvlm,luo2024skysensegpt} have been proposed, introducing a wide range of RS visual knowledge to LVLMs. Along with these datasets, a series of RS-specialized LVLMs \cite{muhtar2024lhrs,kuckreja2024geochat,pang2024h2rsvlm} emerged, achieving markedly improved performance over generic LVLMs on various RS tasks such as visual question answering, visual grounding, scene classification, \etc. 

However, the vision-language understanding granularity of the existing LVLMs in the RS field retains relatively coarse, limited to image-level and region-level. Without the capability of detailed pixel-level interpretation, they can't handle tasks like semantic segmentation. However, fine-grained image understanding is essential for a range of practical applications, including land-cover mapping and environmental protection. 


To address the problems mentioned above, this paper introduces a unified framework that is capable of handling multiple granularity tasks simultaneously. Specifically, we propose RSUniVLM, which is the first RS-specialized VLM that unifies image-level, region-level and pixel-level covering understanding and reasoning tasks, along with multi-image analysis capabilities. As shown in the top left of \cref{fig:teaser}, RSUniVLM preserves visual question answering and captioning at image-level. Besides, it possesses excellent object localization capability at region-level, supporting visual grounding and referring expression generation. Apart from the above capabilities, RSUniVLM extends the RS LVLMs to pixel-level comprehension and multiply images analysis in a unified framework. As shown in the bottom of \cref{fig:teaser}, our RSUniVLM also works well on semantic segmentation, change captioning, and change detection. Specifically, we adopt the semantic descriptor representation method introduced in Text4Seg \cite{lan2024text4seg} to format the segmentation mask, unifying the multi-level perception and generation tasks to text-only generation.

\begin{table}[t!]
\setlength{\tabcolsep}{2.5pt}
  \centering
  \footnotesize
  \begin{tabular}{@{}l|cc|cc|c|cc}
    \toprule
    Method & \multicolumn{2}{c|}{\scriptsize Image-level} & \multicolumn{2}{c|}{\scriptsize Region-level} & \multicolumn{1}{c|}{\scriptsize Pixel-level} &  \multicolumn{2}{c}{\makecell{\scriptsize Multi-Image}} \\
     & SC & VQA & VG & REG  & SS & CC & CD  \\
    \midrule
    LLaVA \cite{liu2024visual}     & \Checkmark & \Checkmark & \XSolidBrush & \XSolidBrush & \XSolidBrush & \XSolidBrush & \XSolidBrush\\
    InternVL-1.5 \cite{chen2024far}  & \Checkmark & \Checkmark & \XSolidBrush & \XSolidBrush & \XSolidBrush & \XSolidBrush & \XSolidBrush\\
    LISA  \cite{lai2024lisa}       & \Checkmark & \Checkmark & \Checkmark & \XSolidBrush & \Checkmark & \XSolidBrush  & \XSolidBrush \\
    NExT-Chat \cite{zhang2023next} & \Checkmark & \Checkmark & \Checkmark & \Checkmark & \Checkmark & \XSolidBrush & \XSolidBrush \\
    LLaVA-Grounding \cite{zhang2025llava}  & \Checkmark & \Checkmark & \Checkmark & \Checkmark & \Checkmark & \XSolidBrush & \XSolidBrush \\
    OMG-LLaVA \cite{zhang2024omg}  & \Checkmark & \Checkmark & \Checkmark & \Checkmark & \Checkmark & \XSolidBrush & \XSolidBrush \\
    LLaVA-OneVision \cite{li2024llava} & \Checkmark & \Checkmark & \XSolidBrush & \XSolidBrush & \XSolidBrush & \Checkmark & \XSolidBrush \\
    MiniCPM-V-2.6 \cite{yao2024minicpm} & \Checkmark & \Checkmark & \XSolidBrush & \XSolidBrush & \XSolidBrush & \Checkmark & \XSolidBrush \\
    \midrule
    RSGPT \cite{hu2023rsgpt} & \Checkmark & \Checkmark & \XSolidBrush & \XSolidBrush & \XSolidBrush & \XSolidBrush & \XSolidBrush \\
    GeoChat \cite{kuckreja2024geochat} & \Checkmark & \Checkmark & \Checkmark & \Checkmark & \XSolidBrush & \XSolidBrush  & \XSolidBrush\\
    LHRS-Bot \cite{muhtar2024lhrs}  & \Checkmark & \Checkmark & \Checkmark & \Checkmark & \XSolidBrush & \XSolidBrush  & \XSolidBrush\\
    CDChat \cite{noman2024cdchat} & \XSolidBrush & \XSolidBrush & \XSolidBrush & \XSolidBrush & \XSolidBrush & \Checkmark  & \XSolidBrush\\
    ChangeChat \cite{deng2024changechat} & \XSolidBrush & \Checkmark & \XSolidBrush & \XSolidBrush & \XSolidBrush & \Checkmark  & \Checkmark\\
    Change-Agent \cite{liu2024change}& \XSolidBrush & \Checkmark & \XSolidBrush & \XSolidBrush & \XSolidBrush & \Checkmark  & \Checkmark\\
    \midrule
    RSUniVLM (Ours)  & \Checkmark & \Checkmark & \Checkmark & \Checkmark & \Checkmark & \Checkmark & \Checkmark\\
    \bottomrule
  \end{tabular}
  \caption{Comparison of capabilities of several representative large VLMs in general and remote sensing domain. Abbreviations SC, VQA, VG, REG, SS, CC, CD in the table stand for scene classification, visual question answering, visual grounding, referring expression generation, semantic segmentation, change captioning and change detection, respectively.}
  \label{tab:capabilities}
\end{table}

In particular, to enhance the model's multi-level comprehension ability, we propose a novel Granularity-oriented Mixture of Experts (G-MoE) architecture to decouple the understanding and reasoning capabilities of LLM at different granularity levels. G-MoE consists of one training-free task router and three experts for image-level, region-level and pixel-level tasks, respectively. Moreover, we collect and reorganize a variety of datasets from both RS-specific domain and general domain, comprising 1.2 million instruction-following data. By applying a two-stage training scheme, RSUniVLM gains powerful multi-level understanding capabilities across various RS tasks (see the radar chart in \cref{fig:teaser}). For the change captioning task, it achieves a competitive performance compared with the best specialized methods on LEVIR-MCI \cite{liu2024change}. On VRSBench-Ref \cite{li2024vrsbench}, RSUniVLM obtains impressive accuracy of 69.31\%, far exceeding the state-of-the-art model GeoChat (19.1\%).

The major contributions of this paper are threefold:

\begin{itemize}
\item We introduce RSUniVLM, a unified end-to-end vision-language model for remote sensing which is capable of multi-level visual understanding. To the best of our knowledge, RSUniVLM is the first RS-specialized VLM that supports a broad variety of tasks, including visual question answering, visual grounding, semantic segmentation, change analysis, \etc.

\item  We design an effective Granularity-oriented Mixture of Experts (G-MOE) architecture to decouple the model's perception capabilities at different visual granularity levels, promoting robust multi-level understanding without increasing computation costs.

\item We conduct extensive experiments across 6 tasks and 13 datasets. RSUniVLM is comparable to some task-specific models, and achieves state-of-the-art performance on the visual grounding task. Along with its excellent performance, our RSUniVLM only consists of 1B parameters, significantly smaller than many existing LVLMs.
\end{itemize}

\begin{figure*}
    \centering
    \includegraphics[width=1\linewidth]{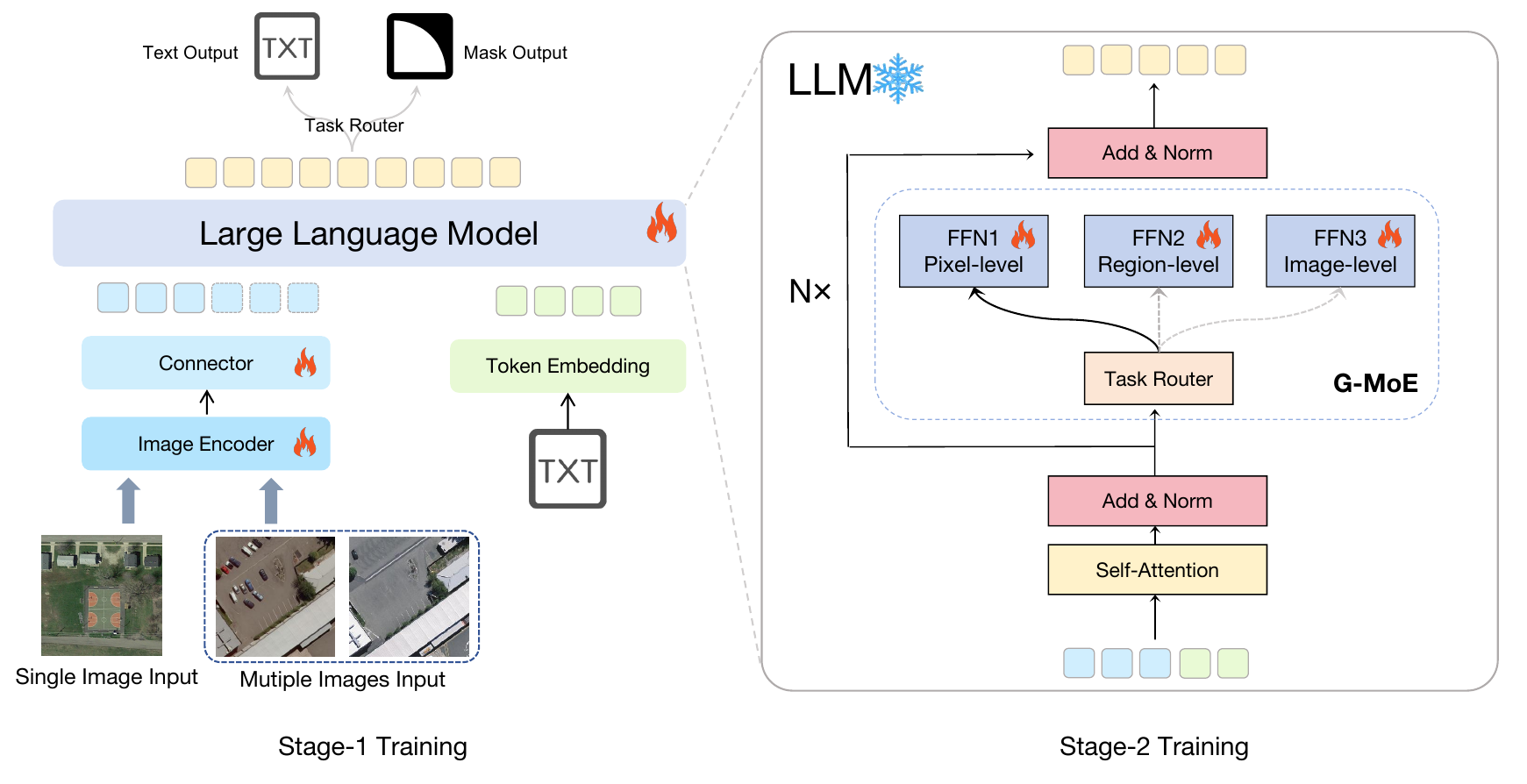}
    \caption{An overview of RSUniVLM. We adopt the classic LLaVA-based architecture, which consists of an image encoder, an MLP connector and a large language model. We unify all tasks into text-only generation, enabling joint optimization for multiple tasks in an end-to-end manner. During stage-2 training, Granularity-oriented MoE is applied to the LLM to enhance model's multi-level vision understanding.}
    \label{fig:model}
\end{figure*}
\section{Related Work}
\label{sec:relatedwork}

\subsection{Large Vision-Language Models.}
In recent years, Large Vision-Language Models (LVLMs) \cite{liu2024visual,zhu2023minigpt,alayrac2022flamingo,chen2024far,bai2023qwen} have made significant advancements by extending the capability of Large Language Models (LLMs) \cite{radford2018improving,radford2019language,guu2020retrieval,raffel2020exploring} to multi-modal understanding and reasoning via vision perception modules. Initial research primarily focuses on vision-language reasoning and comprehension. For instance, a series of works, like Flamingo \cite{alayrac2022flamingo}, LLaVA \cite{liu2024visual}, Instruct-BLIP \cite{instructblip} and MiniGPT-4 \cite{zhu2023minigpt} have shown impressive visual question answering and multimodal chat abilities. VideoChat \cite{li2023videochat}, LLaMA-VID \cite{li2025llama} and Qwen-Audio \cite{chu2023qwen} expand the capabilities to video-text and audio-text understanding. Moreover, recent works like UnifiedMLLM \cite{li2024unifiedmllm}, VITRON \cite{fei2024vitron} and VisionLLM v2 \cite{wu2024visionllm} explore universal modalities of inputs and outputs through a unified representation, supporting a broad variety of tasks such as segmentation, image generation and video generation. These models fully demonstrate the powerful ability of LVLMs in visual perception and reasoning.

\subsection{Remote Sensing Vision-Language Models.}
Accompanying the rapid development of generic LVLMs, advancements in remote sensing (RS) LVLMs \cite{hu2023rsgpt,muhtar2024lhrs,kuckreja2024geochat,pang2024h2rsvlm} have alse achieved promising progress within the two years. Inspired by MiniGPT-4 \cite{zhu2023minigpt}, RSGPT \cite{hu2023rsgpt} is proposed to align visual features of RS images with LLMs, supporting image captioning and visual question answering tasks. GeoChat \cite{kuckreja2024geochat} achieves the ability of visual grounding using a novel RS multimodal instruction following dataset. LHRS-Bot \cite{muhtar2024lhrs} constructs a large-scale instruction dataset, LHRS-Instruct, leveraging the globally available RS images and the corresponding features from OpenStreetMap. Based on LHRS-Instruct, LHRS-Bot exhibits a profound comprehension of RS image. SkyEyeGPT \cite{zhan2024skyeyegpt} is first to support remote sensing video captioning task. In addition, there are also some models specifically designed for the RS image change interpretation task based on LVLMs, such as Change-Agent \cite{liu2024change}, CDChat \cite{noman2024cdchat}, ChangeChat \cite{deng2024changechat}.

Although these RS-specialized LVLMs have developed powerful capabilities in RS image understanding and can handle multifarious tasks, their abilities are still relatively limited compared to the unified LVLMs in general domain. Our RSUniVLM is the first unified RS VLM model which achieves image-level, region-level, and pixel-level reasoning and perception ability in an end-to-end manner. 

\subsection{Mixture of Experts in Multi-modal Learning.}
To effectively scale model parameters without a corresponding increase in computational demand, the mixture of experts (MoE) architecture has emerged as a possible solution. MoE mainly consists of two components: a set of expert layers and a gated router. The feed
forward blocks are replaced by the expert layers, and the router determines the token set that each expert handles, which reduces interference between different types of inputs. MoE has maintained a robust trajectory of growth, particularly notable with the advent of Mixtral-8x7B \cite{jiang2024mixtral} and a variety of subsequent industrial-scale LLMs such as LLaVA-MoE \cite{lin2024moe}, DBRX \cite{gupta2024dbrx}, DeepSeek-V2 \cite{liu2024deepseek}. In this work, we propose a new sparse Granularity-oriented MoE architecture, enhancing the capability of multi-modal understanding.

\section{Method Description}
\subsection{Model Design}

The architecture of RSUniVLM is illustrated in \cref{fig:model}. Following the design paradigm of commonly used LLaVA-style frameworks, the proposed RSUniVLM consists of the following four key components: (1) an image encoder pre-trained on extensive image-text datasets; (2) a word embedding layer for text embedding extraction; (3) a multi-layer projector (MLP) that facilitates alignment between image tokens and text embedding; and (4) a large language model (LLM) that jointly processes the aligned image and text embedding to generate textual outputs in an auto-regressive manner. For the input with multiple images, we use a weight-shared image encoder to extract features from each image separately, then we concatenate them along the embedding dimension directly.

\paragraph{Unified Representation.} We formulate all tasks to text-only generation, including object localization and segmentation. For region-specific tasks such as visual grounding and referring expression generation, all bounding boxes are normalized to integer values in the range of $[0, 100]$ and represented as the textual format: [$x_1, y_1, x_2, y_2$]. As for mask generation tasks such as semantic segmentation and change captioning, we adopt the semantic descriptors representation method proposed in \cite{lan2024text4seg}. Specifically, we encode a mask into a sequence of semantic descriptors, which enables LLM to gain the ability of mask generation. The unified representation eliminates additional task heads, making it easy to conduct multi-task joint optimization and leverage large-scale instruction following data. During the inference stage, the textual response of the LLM will be transponded based on the type of task. 

\begin{table*}
  \centering
  \begin{tabular}{@{}l|cccc|cccc}
    \toprule
     Method & \multicolumn{4}{c|}{RSVQA-LR}  & \multicolumn{3}{c}{RSVQA-HR} \\
     & Rural/Urban &  Presence &  Compare &  Avg. & Presence &  Compare &  Avg. \\
    \midrule
    LLaVA-1.5 \cite{liu2024improved} &  59.22 &  73.16 &  65.19 &  65.86 &  48.96 &  59.02 &  53.99 \\
    MiniGPTv2 \cite{zhu2023minigpt}   &  60.02  &  51.64  &  67.64 &   59.77 &   68.34 &   64.71 &   66.53 \\
    QWen-VL-Chat \cite{bai2023qwen}   &  62.00 &   47.65  &  66.54 &   58.73 &   61.75 &   65.98 &   63.87 \\
    InternLM-XCompose \cite{zhang2024internlm}&   59.00  &  66.74 &   52.91  &  59.55  &  67.79 &   66.62 &   67.21 \\
    GeoChat-7B \cite{kuckreja2024geochat} & 91.09 &  90.33 &\textbf{94.00}&  91.81 &  58.45  & 83.19 &  70.82 \\
    SkyEyeGPT-7B \cite{zhan2024skyeyegpt} & 88.93  & 88.63  & 75.00  & 84.16 &  80.00 &  80.13 &  82.56\\
    LHRS-Bot-7B \cite{muhtar2024lhrs} & 89.07 &  88.51 &  90.00  & 89.19 & \textbf{ 92.57} & \textbf{ 92.53 } & \textbf{92.55} \\
    H2RSVLM-7B \cite{pang2024h2rsvlm} & 88.00 &  89.58 &  89.79  & 89.12  & 65.00 &  83.70 &  74.35 \\
    \bottomrule
    RSUniVLM-1B & \textbf{92.00} &\textbf{ 91.51} & 92.65 & \textbf{92.05 }& 90.81 & 90.88 & 90.85 \\
    \bottomrule
  \end{tabular}
  \caption{Accuracy of different models on visual question answering datasets RSVQA-LR and RSVQA-HR \cite{lobry2020rsvqa}. Following the practice of prior works \cite{hu2023rsgpt,kuckreja2024geochat,muhtar2024lhrs}, we omit area and count questions during evaluation.}
  \label{tab:vqa}
\end{table*}

\begin{table}
\setlength{\tabcolsep}{5pt}
  \centering
  \begin{tabular}{@{}l|cp{1cm}cp{1cm}ccc}
    \toprule
    Method &  AID & WHU- \newline RS19  &  NWPU &  SIRI-WHU  \\
    \midrule
    LLaVA-1.5 \cite{liu2024improved} &  31.10  & 54.55 &  34.96 &  17.71   \\
    MiniGPTv2 \cite{zhu2023minigpt} &  32.96  & 64.80  & 28.15  & 35.46  \\
    InstructBLIP \cite{instructblip} &  29.50  & 36.76 &  34.01  & 18.20  \\
    QWen-VL-Chat \cite{bai2023qwen} &  55.30 &  72.25 &  42.73 &  54.58 \\
    GeoChat-7B \cite{kuckreja2024geochat} & 72.03 & 82.33  & -  &  43.67  \\
    LHRS-Bot-7B \cite{muhtar2024lhrs} & \textbf{ 91.26} &  \textbf{93.17} &  83.94  & 62.66 \\
    \bottomrule
    RSUniVLM-1B & 81.18 & 84.91  & \textbf{ 86.86} & \textbf{68.13} \\
    \bottomrule
  \end{tabular}
  \caption{Scene Classification accuracy of different models.}
  \label{tab:cls}
\end{table}

\subsection{Granularity-oriented Mixture of Experts}

Motivated by the existing mixture-of-experts methods, we design a novel model architecture: Granularity-oriented Mixture of Experts (G-MoE), as shown in \cref{fig:model}. G-MoE decouples experts based on three different granularities of task type:

\begin{itemize}
\item \textit{Image-level Expert}. This expert operates at the holistic image level, capturing high-level visual information and global semantics across the entire image. It's particularly beneficial for tasks like image captioning and change captioning by extract general contextual information.

\item \textit{Region-level Expert}. Focused on distinct regions within the image, this expert specializes in identifying and understanding localized patterns that are crucial for more detailed analysis, such as object localization or referring expression generation.

\item \textit{Pixel-level Expert}. Operating at the finest level, this expert focuses on comprehending semantic information at the pixel level. The fine-granularity perception is essential for tasks that require high accuracy in detecting intricate details, such as fine-grained change detection or semantic segmentation.  
\end{itemize}
To integrate these experts effectively, a training-free gating mechanism is employed. This mechanism assigns an input prompt to the specific expert based on the characteristics of the data, ensuring that the model's response is both context-aware and efficient.

\subsection{Training Strategy}

We adopt a two-stage coarse-to-fine training strategy: an initial multi-task pre-training stage, followed by a fine-tuning stage to further improve visual understanding and reasoning across multiple levels.

\paragraph{Stage 1.} In the first stage, we aim to inject domain-specific knowledge from the remote sensing field into a pre-trained VLM, and enable the model to acquire initial multi-task processing capabilities by applying full-parameter fine-tuning. To create a robust and instruction-following dataset, we integrate fifteen diverse public datasets covering five distinct tasks in the remote sensing field, then we convert them into structured instruction-following sets with manually crafted templates. Additionally, we incorporate a part of several high-quality instruction sets from RS and general domain. \cref{subsec:ExperimentalSetup} shows the details of the dataset. During this stage, the Granularity-oriented Mixture-of-Experts (G-MoE) layers have not been introduced to the LLM yet, enabling the model to focus on foundational alignment.

\paragraph{Stage 2.}  In the second stage, we enhance the model's specialization by duplicating the feed-forward network (FFN) layers three times to initialize dedicated experts for different types of remote sensing tasks. This makes each expert be able to better capture unique task-specific patterns, thereby refining the model's performance across diverse applications. We sample a small portion of RS-specific instruction data from the stage 1's training set based on the granularity and variety of tasks, to further fine-tune the G-MoE layers.

The objective of these two training stages is identical. For a sequence of length $L$, we compute the probability of the target answers $X_a$ by: 
\begin{equation}
  p(X_a|X_v,X_{instruct}) = \prod_{i=1}^{L}p_\theta(x_i|X_v,X_{instruct,<i},X_{a<i})
  \label{eq:objective}
\end{equation}
where $\theta$ is the trainable parameters, $X_{instruct,<i}$ and $X_{a,<i}$ are the instruction and answer tokens in all turns before the current prediction token $x_i$ , respectively. $X_v$ represents the aligned tokens of the input image, and for a multi-image input scene, we directly concatenate the image tokens.
\begin{table}
 \setlength{\tabcolsep}{3.5pt}
  \centering
  \begin{tabular}{@{}l|>{\centering\arraybackslash}p{1.05cm}>{\centering\arraybackslash}p{1.05cm}|>{\centering\arraybackslash}p{1.05cm}>{\centering\arraybackslash}p{1.05cm}}
    \toprule
    Method & \multicolumn{2}{c|}{DIOR-RSVG} & \multicolumn{2}{c}{VRSBench-Ref}\\
    & {\small acc@0.5} & {\small acc@0.7} & {\small acc@0.5} & {\small acc@0.7} \\
    \midrule
    CogVLM*-17B \cite{wang2023cogvlm}& 44.50 & 30.2 & 20.31 & 15.22 \\
    Florence2-0.7B \cite{xiao2024florence}& 18.95 & 11.71 & 21.69 & 12.74 \\
    GeoChat-7B \cite{kuckreja2024geochat} & 27.85 & 9.50 & 39.6 & 19.1 \\
    LHRS-Bot-7B \cite{muhtar2024lhrs} & 12.95 & 8.95 &  1.74 & 0.17 \\
    H2RSVLM-7B \cite{pang2024h2rsvlm} & 48.04 & - &  - & - \\
    \bottomrule
    RSUniVLM-1B & \textbf{72.47} &\textbf{ 56.17} &\textbf{ 69.31} &\textbf{ 47.47} \\
    \bottomrule
  \end{tabular}
  \caption{Visual Grounding performance on the DIOR-RSVG \cite{zhan2023rsvg} and VRSBench \cite{li2024vrsbench} datasets. Accuracy is calculated with the threshold of 0.5 and 0.7 respectively. CogVLM* indicates the CogVLM-grounding-generalist version. The results for LHRS-Bot are re-evaluated using its updated evaluation code.}
  \label{tab:vg}
\end{table}

\section{Experiments}

\begin{table*}
  \centering
  \begin{tabular}{@{}l|cccccccccc}
    \toprule
    Method &  BLEU-1 &  BLEU-2 &  BLEU-3 &  BLEU-4 &  METEOR &  ROUGE-L &  CIDEr \\
    \midrule
    InternVL2-8B \cite{chen2023internvl} & 25.28 & 10.74 & 3.90 & 1.19 & 16.46 & 17.11 & 3.04 \\
    MiniCPMV2.6-17B \cite{yao2024minicpm} & 17.27 & 9.08 & 4.09 & 1.70 & 15.07 & 18.85 & 1.85 \\
    RSICCFormer \cite{liu2022remote} &  83.09 &  74.32  & 66.66 &  60.44  & 38.76  & 72.63  & 130.00\\
    Prompt-CC \cite{liu2023decoupling} &  83.66 &  75.73 &  69.10 &  63.54  & 38.82  & 73.72 &  136.44 \\
    Sen \cite{zhou2024single} &  85.10  & 77.05 &  70.01 &  64.09  & 39.59 &  74.57 &  136.02 \\
    SFT \cite{sun2024lightweight} &  84.56 &  75.87 &  68.64 &  62.87  & 39.93  & 74.69  & 137.05 \\
    Chg2Cap \cite{chang2023changes} &  86.14  & \textbf{78.08 } & 70.66 & \textbf{ 64.39} &  40.03 &  75.12  & 136.61 \\
    RsCaMa \cite{liu2024rscama} &  85.79  & 77.99  & \textbf{71.04} &  65.24 &  39.91  & 75.24 &  136.56 \\
    \bottomrule
    RSUniVLM &\textbf{ 86.30} & 75.86  &  67.49 & 60.27  & \textbf{ 40.14 }&\textbf{ 80.09} & \textbf{139.80}\\
    \bottomrule
  \end{tabular}
  \caption{Change Captioning performance comparison on LEVIR-MCI \cite{liu2024change}. we employ BLEU for precision between hypotheses and reference sentences, METEOR for alignment consideration of synonyms and word order, ROUGE-L for recall based on Common Subsequence, and CIDEr for similarity with multiple references. }
  \label{tab:cc}
\end{table*}

\begin{table}
  \setlength{\tabcolsep}{4pt}
  \centering
  \begin{tabular}{@{}l|cccccccc}
    \toprule
    Method & F1 & Pre. & Rec. & OA  & IoU \\
    \midrule
    AdvNet \cite{vu2019advent} & 71.10&78.76&64.80&97.91& 55.16\\
    S4GAN \cite{mittal2019semi} & 73.39 & 81.56 & 66.70 & 98.08 &  57.96 \\
    SemiCDNet \cite{peng2020semicdnet} & 70.07 & 79.97 & 62.36 & 97.89 & 53.93 \\
    SemiCD \cite{bandara2022revisiting} & \textbf{78.76 }&\textbf{ 86.22} &\textbf{ 72.49 }&\textbf{ 98.45} &\textbf{64.96}\\
    RCL \cite{wang2022reliable} & 71.56 & 74.63 & 68.73 & 97.83 & 55.71\\
    \bottomrule
    RSUniVLM & 71.38 & 73.13 & 67.80 & 92.64 & 54.19 \\
    \bottomrule
  \end{tabular}
  \caption{Quantitative  comparison on WHU-CD \cite{ji2018fully}. Compared with the specialized models trained with 5\% labeled samples, RSUniVLM performs competitively in a zero-shot setting.}
  \label{tab:CD}
\end{table}

In this section, we begin by detailing the composition of the instruction datasets used in the two-stage training scheme. Next, we present quantitative comparisons of our model with others across five tasks, followed by the qualitative results. Finally, we provide ablation study results to demonstrate the effectiveness of our G-MoE architecture.

\subsection{Experimental Setup}
\label{subsec:ExperimentalSetup}
\paragraph{Dataset Details.} To equip our model with perception and understanding abilities across various levels of visual granularity, we collect and reorganize a large-scale instruction fine-tuning dataset, sourced from a wide range of domains and encompassing multiple task types. For image-level tasks, we sample from datasets across three tasks: Visual Question Answering, scene classification, and image captioning, including RSVQA \cite{lobry2020rsvqa}, NWPU \cite{cheng2017remote}, fMoW \cite{fmow2018}, RSTeller \cite{ge2024rsteller}, \etc. For region-level tasks, we utilize two prominent open-access visual grounding datasets, DIOR-RSVG \cite{zhan2023rsvg} and OPT-RSVG \cite{li2024language}. For the pixel-level task, we reorganize several RS semantic segmentation datasets, \eg , landcover \cite{boguszewski2021landcover}, OpenEarthMap \cite{xia2023openearthmap}, Potsdam\footnote{https://www.isprs.org/education/benchmarks/UrbanSemLab\label{web}}, together with the COCO stuff \cite{caesar2018coco} dataset. We represent each mask with $24\times24$ semantic descriptors by row-wise RLE proposed in \cite{lan2024text4seg}. Besides, we include LEVIR-MCI \cite{liu2024change} to improve multi-image comprehension. Finally, we incorporate a substantial amount of high-quality instruction data in both RS and general domains, including GeoChat-instruction \cite{kuckreja2024geochat}, VRSBench \cite{li2024vrsbench}, ShareGPT4V \cite{chen2023sharegpt4v}, All-SeeingV2 \cite{wang2025all}, \etc to enhance the conversation and understanding ability. We obtain a total of 1.2M instruction data for the first stage training. Next, in the second training stage, we remove the general instruction datasets and perform dataset resampling for each remote sensing task separately. We elaborate on the details of datasets in the supplementary.  

\paragraph{Implementation Details.} We adopt the pre-trained SigLIP-400m as the image encoder, QWen2-0.5B as the LLM and 2-layer MLPs as the vision-language connector. Moreover, we add a unique task identifier for each task type as in previous work \cite{kuckreja2024geochat,muhtar2024lhrs}. The number of granularity-oriented experts is set to 3, corresponding to three different levels. For the first training stage, we adopt the full-parameter fine-tuning method leveraging the entire dataset for 1 epoch. At the second stage, we only fine-tune G-MoE layers in LLM using datasets of three visual granularity levels. The AdamW optimizer are used with the initial learning rate of 1e-5, a warmup ratio of 0.03, and a batch size of 4. All models are trained on 4 Nvidia A40 GPUs (40GB) for approximately 30 hours in total. More training details are provided in the supplementary material.

\begin{table}
  \setlength{\tabcolsep}{4pt}
  \centering
  \begin{tabular}{@{}l|cccccc}
    \toprule
    Method & Vaihingen & UDD5 & VDD & Avg.  \\
    \midrule
    MaskCLIP \cite{zhou2022extract} &  24.7 &  32.4 &  32.9 & 30.0 \\
    SCLIP \cite{wang2025sclip}&  28.4 &  38.7 &  37.9 & 35.0 \\
    GEM \cite{bousselham2024grounding} &  24.7 &  41.2 & 39.5 & 35.13 \\
    ClearCLIP \cite{lan2024clearclip} &  27.3 & 41.8 &  39.3 & 36.13 \\
    SegEarth-OV \cite{li2024segearth} & 29.1 & 50.6  & 45.3 & 41.67 \\
    \bottomrule
    RSUniVLM & \textbf{54.46} & \textbf{65.41} & \textbf{49.6} & \textbf{56.49} \\
    \bottomrule
  \end{tabular}
  \caption{Zero-Shot Semantic Segmentation quantitative comparison on Vaihingen, UDD5 \cite{chen2018large} and VDD \cite{cai2305vdd}. The evaluation metrics is mean intersection over union (mIoU).}
  \label{tab:seg}
\end{table}

\begin{table}
  \setlength{\tabcolsep}{4pt}
  \centering
  \begin{tabular}{@{}l|cccccccccc}
    \toprule
    Method & \makecell{Trainable \\ Parameters } &  \makecell{Runtime \\ Parameters } &VQA & VG & \\
    \midrule
    LoRA & 72M & 987M & 90.80 & 64.56 \\
    MoE & 793M & 779M & 82.75 & 49.44\\
    Full-parameter & 893M & 893M & 90.22 & 65.27 \\
    G-MoE & 313M & 893M & \textbf{91.57 }& \textbf{70.90} & \\
    \bottomrule
  \end{tabular}
  \caption{Performance comparison using different fine-tuning methods and applying different MoE architecture.}
  \label{tab:ablation}
\end{table}

\begin{figure*}
    \centering
    \includegraphics[width=\textwidth]{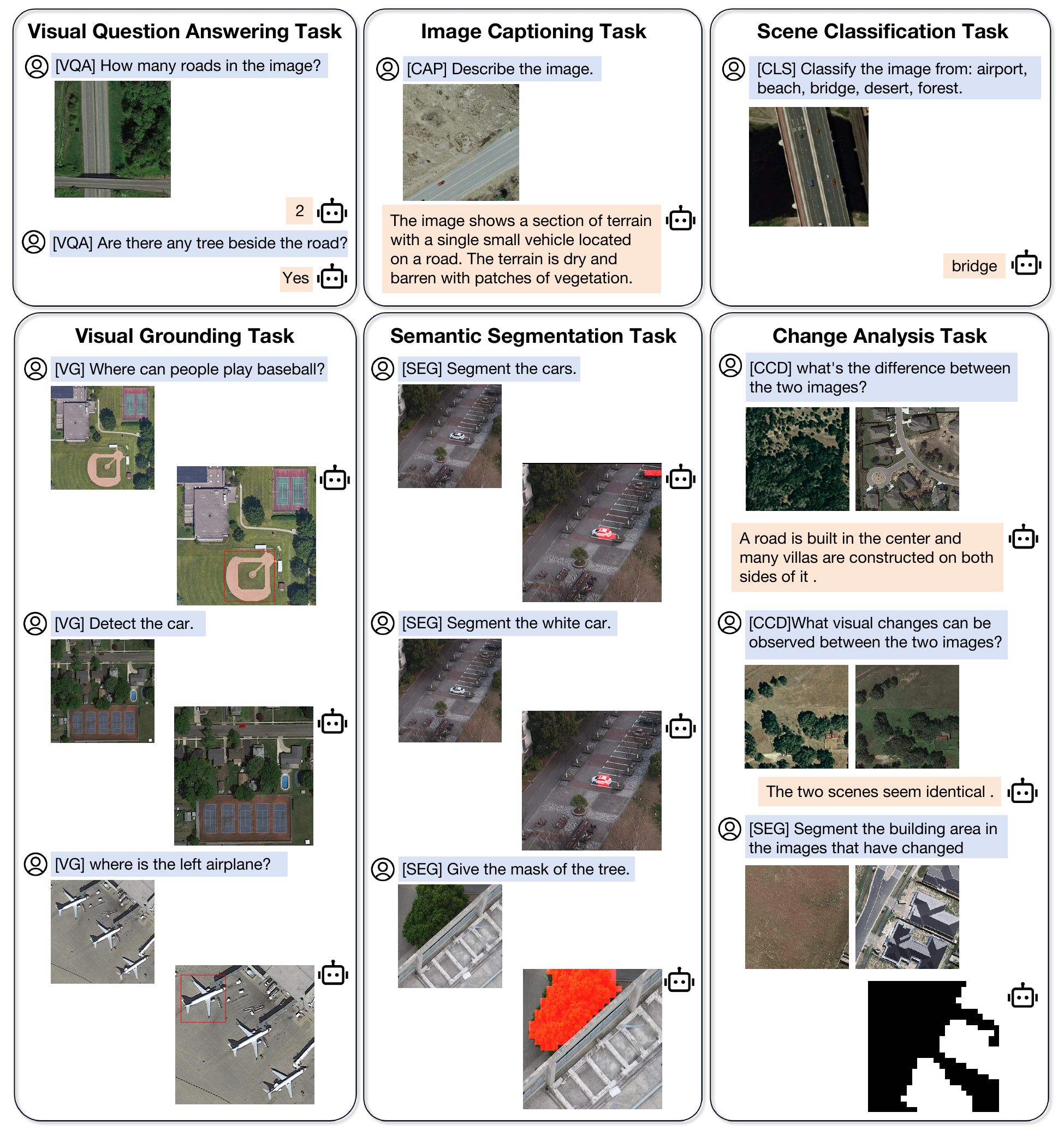}
    \caption{Qualitative Results of RSUniVLM across a variety of tasks, demonstrating the ability of RSUniVLM to handle multi-level visual granularity tasks. Moreover, RSUniVLM perform well in change analysis involving multi-image input. }
    \label{fig:Qualitative}
\end{figure*}

\subsection{Quantitative Results}

\paragraph{Scene Classification.}
We evaluate our model on four RS scene classification datasets: AID \cite{xia2017aid}, WHU-RS19 \cite{dai2010satellite}, NWPU-RESISC45 \cite{cheng2017remote}, SIRI-WHU \cite{zhu2016bag}. We test on the datasets in a zero-shot setting except for NWPU. The classification results are presented in \cref{tab:cls}, from which we can see that our RSUniVLM achieves best performance on SIRI-WHU and NWPU compared with other models, with an accuracy of 86.86\% and 68.13\%, respectively.

\paragraph{Visual Question Answering.}
We evaluate our model's VQA performance on RSVQA-LR and RSVQA-HR \cite{lobry2020rsvqa}. As shown in \cref{tab:vqa}, RSUniVLM exhibits robust performance across both high-resolution and low-resolution settings, and especially excels at LR setting, with a much smaller model size (1B) compared to others (e.g., 7B).

\paragraph{Visual Grounding.}
We use DIOR-RSVG \cite{zhan2023rsvg} and VRSBench-Ref \cite{li2024vrsbench} for the evaluation of visual grounding. The results are shown in \cref{tab:vg}, from which we can see that the proposed RSUniVLM has outstanding fine-grained object localization ability in both the general and RS fields, achieving an accuracy of 69.31\% on VRSBench-Ref, which is significantly better than geochat's 39.6\%.

\paragraph{Change Captioning \& Change Detection.} \cref{tab:cc} presents the results of change captioning on LEVIR-MCI \cite{liu2024change}. Different from the existing RS VLMs such as GeoChat and LHRS-Bot, that are unable to support multi-image reasoning, RSUniVLM is capable of comprehending semantic level interpretation information of changes observed in RS images. Experimental results show that our RSUniVLM significantly outperforms the generic models and is comparable to state-of-the-art task-specific models. 
For the change detection task, we evaluate RSUniVLM's zero-shot performance on WHU-CD \cite{ji2018fully}. The results are presented in the \cref{tab:CD}. During the evaluating stage, we apply the Chain-of-Thought \cite{wei2022chain} technology to improve the performance. It can be observed that our model performs competitively in a zero-shot setting, while all the other models use 5\% labeled data for supervision.

\paragraph{Zero-Shot Semantic Segmentation.}
We evaluate the zero-shot semantic segmentation performance on Vaihingen\textsuperscript{\ref {web}}, UDD5 \cite{chen2018large} and VDD \cite{cai2305vdd} using the mean intersection over union (mIoU), compared with other Clip-based open-vocabulary segmentation methods, such as MaskCLIP \cite{zhou2022extract}, ClearCLIP \cite{lan2024clearclip}, \etc. As shown in \cref{tab:seg}, our RSUniVLM achieves a significant improvement across all datasets, with an average gain of 35\% compared to SegEarth-OV \cite{li2024segearth}.

\subsection{Qualitative Results}
We provide some examples of qualitative results in \cref{fig:Qualitative}, which clearly demonstrate the broad capabilities of our RSUniVLM. The figure displays 6 instances across various tasks: image-level visual question answering; region-level visual grounding; pixel-level semantic segmentation; change analysis with multiple image inputs. To the best of our knowledge, using our RSUniVLM, for the first time, these diverse RS tasks in various granularities can be effectively handled in one single unified model.

\subsection{Ablation Analysis} 
We evaluate the effectiveness of different adaptation methods, including Low-Rank Adaptation (LoRA), Mixture of Experts (MoE), full-parameter fine-tuning and our Granularity-oriented Mixture of Experts. For LoRA, we set the rank r and the scale factor $\alpha$ to 128 and 256, respectively. For the basic MoE, we set the numbers of experts to 8. As for training, We employ full-parameter fine-tuning on a pre-trained model using the full dataset mentioned in \cref{subsec:ExperimentalSetup}. Then, we apply these four adaptation methods to fine-tune the model across our stage-2 datasets for 1 epoch.

For the VQA task, we calculate the average results on RSVQA-LR and RSVQA-HR. For the VG task, we calculate the average accuracy with the threshold of 0.5 on DIOR-RSVG and VRSBench-Ref. The results are presented in \cref{tab:ablation}. Compared with LoRA and full-parameter fine-tuning methods, G-MoE maintains a moderate level of trainable and runtime parameters while achieving the best performance; compared with the classic MoE architecture, G-MoE improves the VQA performance by an average of 10\% and the VG performance by an average of 40\%, with almost no runtime computational cost increasement.

\section{Limitation} 
Although RSUniVLM is capable of handling various remote sensing tasks effectively and efficiently, its multi-turn conversations capability is relatively weak. This could be improved in the future by involving more high-quality multi-turn conversations data. Furthermore, RSUniVLM is unable to perform generative tasks in the RS field, such as remote sensing image super-resolution and dehazing.

\section{Conclusion} 
In this paper, we presented RSUniVLM, the first unified remote sensing VLM that exhibits the ability of handling diverse RS tasks, including visual question answering, visual grounding, change captioning, change detection, and semantic segmentation, \etc. By adopting a unified text-only representation, RSUniVLM is fully end-to-end trainable, eliminating the need for additional task-specific heads. Moreover, we proposed a novel Granularity-oriented Mixture of Experts architecture and employ a corresponding two-stage training scheme, enabling our model to achieve impressive performance in various tasks, with significant reduction of trainable parameters and computation costs.

{
    \small
    \bibliographystyle{ieeenat_fullname}
    \bibliography{main}
}

\clearpage
\setcounter{page}{1}
\maketitlesupplementary

\begin{figure*}
    \centering
    \includegraphics[width=\textwidth]{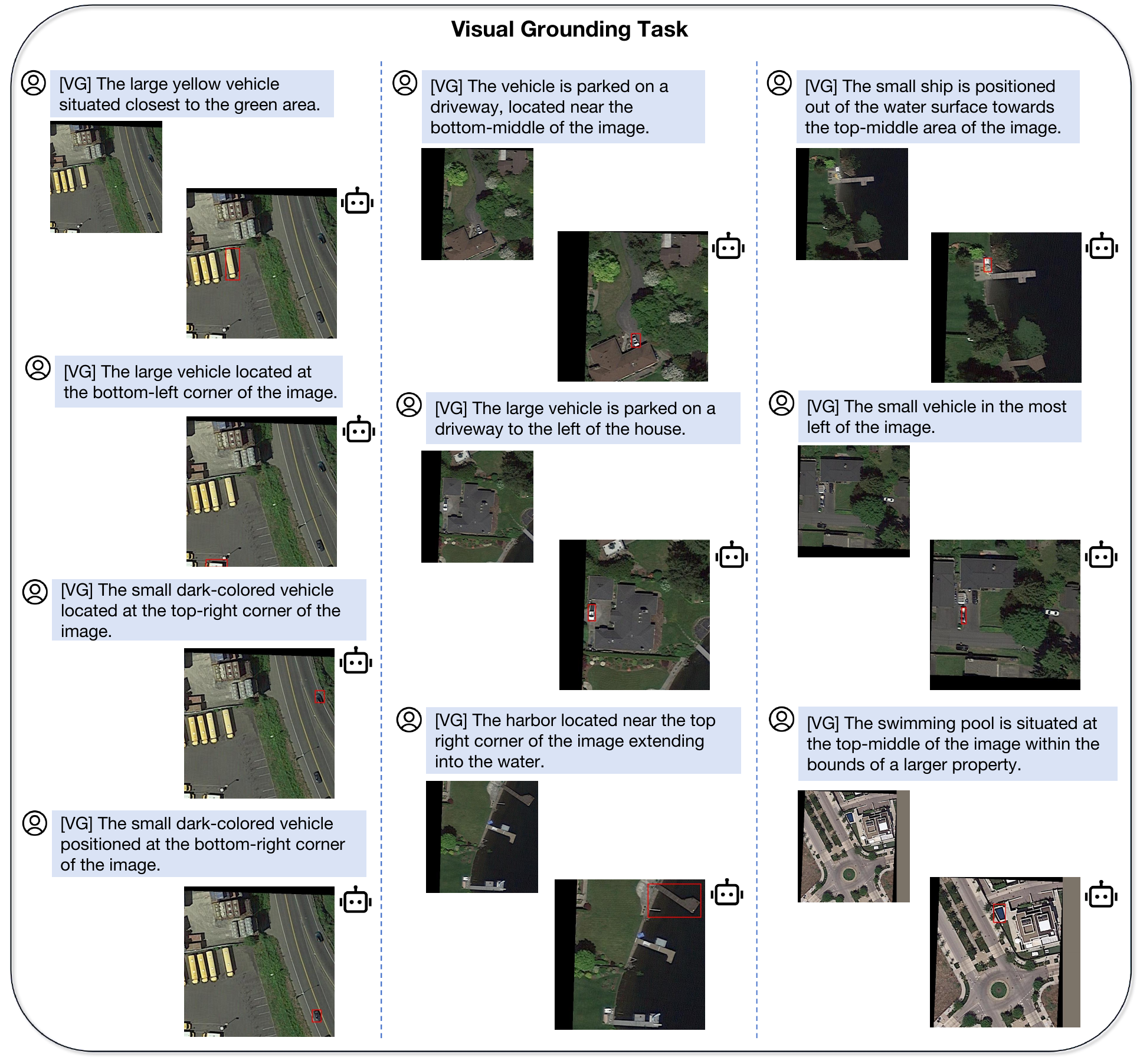}
    \caption{Qualitative Results of visual grounding. }
    \label{fig:suppl_vg}
\end{figure*}

\begin{figure*}
    \centering
    \includegraphics[width=\textwidth]{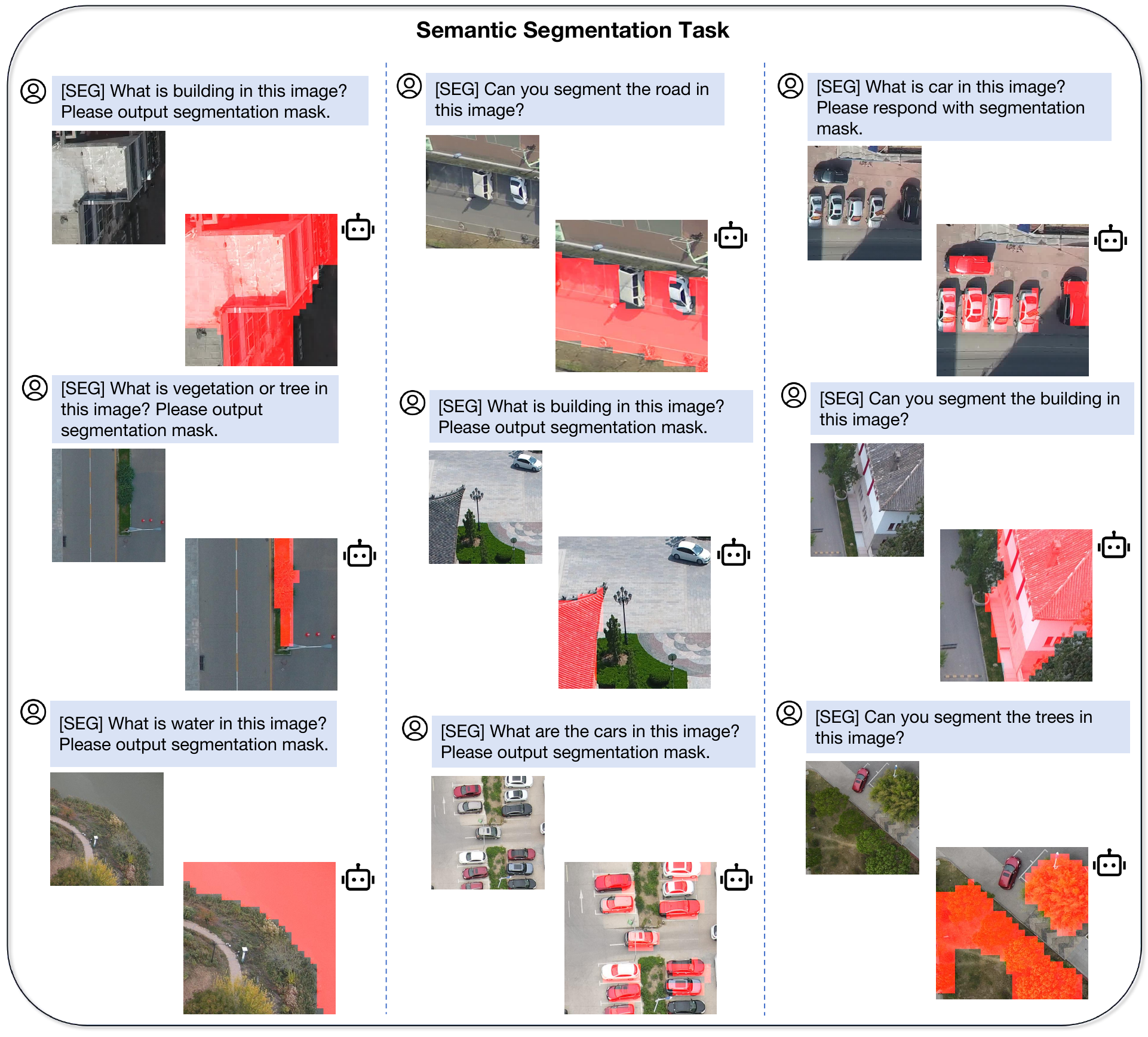}
    \caption{Qualitative Results of semantic segmentation. }
    \label{fig:suppl_seg}
\end{figure*}

\begin{figure*}
    \centering
    \includegraphics[width=\textwidth]{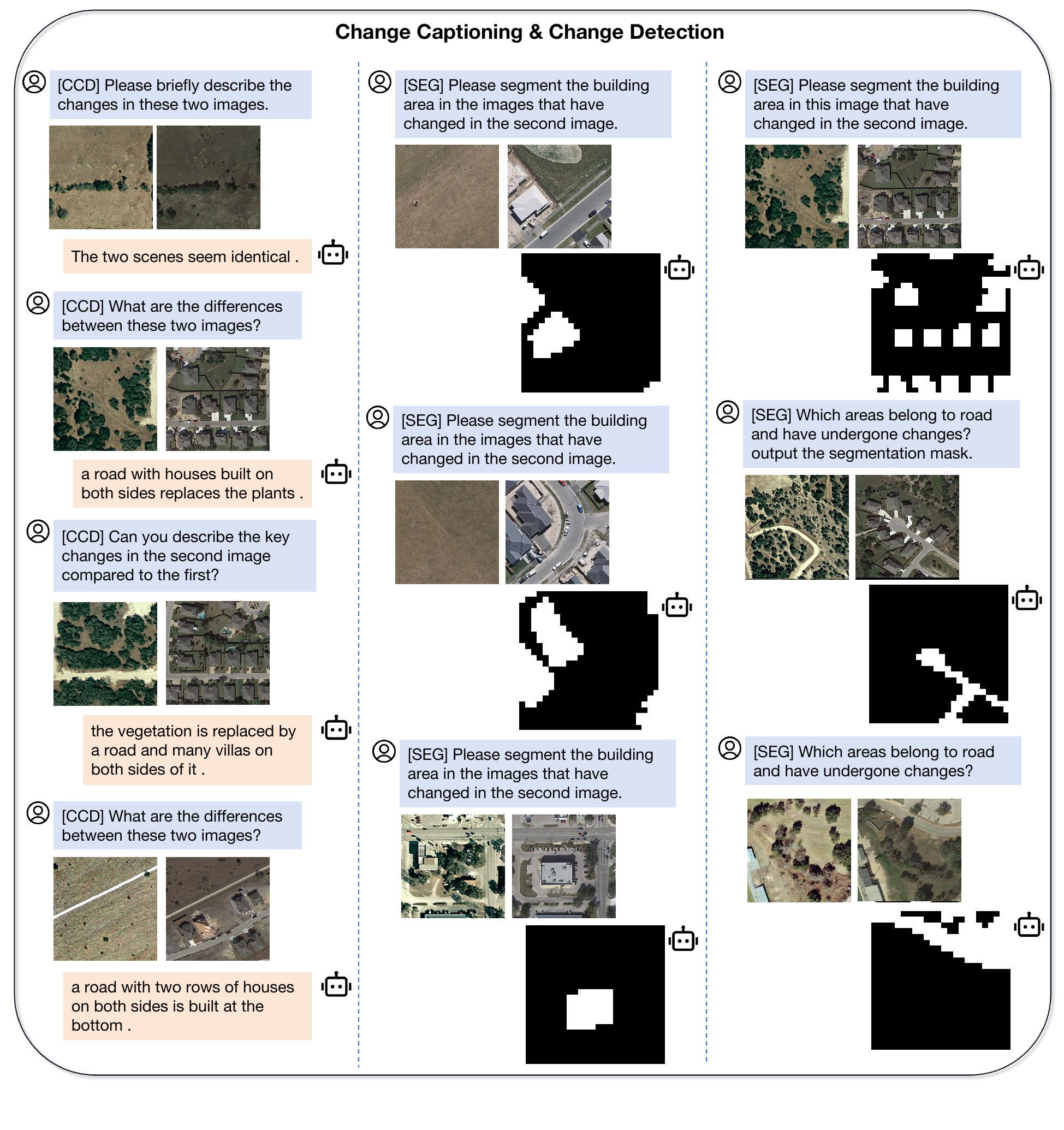}
    \caption{Qualitative Results of change captioning and change detection. }
    \label{fig:suppl_ccd}
\end{figure*}

\section{Details of Dataset}
\label{sec:Dataset}
We construct a universal multi-task instruction-following dataset by leveraging a variety of public RS datasets, together with several existing RS and general instruction datasets. As shown in \cref{tab:multi-taskRS}, we collect and reorganize 15 public datasets across 6 tasks, and we leverage 3 high-quality RS instruction-following datasets: GeoChat \cite{kuckreja2024geochat}, VRSBench \cite{li2024vrsbench} and FIT-RS \cite{luo2024skysensegpt}. Moreover, we found that incorporating general domain instruction datasets into training data enhances multi-modal reasoning and understanding. Therefore, we make use of portions of 6 general-domain instruction datasets as shown in \cref{tab:general instruction}. Besides, COCO stuff is utilized to improve model's performance of semantic segmentation. We use all the aforementioned data for the first stage of model training. For the segmentation task, we divide the mask into $24\times24$ semantic descriptors following \cite{lan2024text4seg}.

For the second stage of training, we exclusively use data from the remote sensing domain, which is divided into three parts based on granularity: semantic segmentation and change detection data for pixel-level, visual grounding and referring expression generation data for region-level, and the remainder for image-level. The contents of the division datasets are detailed in \cref{tab:stage2data}.

\section{Task Tokens}
\label{sec:token}
To enable RSUniVLM to distinguish different tasks more flexibly, we insert a task-specific token at the beginning of the text prompt, following the previous works \cite{kuckreja2024geochat,muhtar2024lhrs}. Specially, we add token [CAP], [CLS], [VQA], [VG], [REF], [SEG] and [CCD] for image captioning, scene classification, visual question answering, visual grounding, referring expression generation, semantic segmentation and change analysis, respectively.

\begin{table}
  \centering
  \begin{tabular}{@{}l|cccccccccc}
    \toprule
     &  Stage1 & Stage2\\
    \midrule
    Trainable Parts & \makecell{Image Encoder, \\ Connector, LLM }& G-MoE \\
    Optimizer & AdamW & AdamW \\
    Learning Rate & 1e-5 & 1e-5\\
    Weight Decay & 0. & 0. \\
    Warmup Ratio & 0.03 & 0.03 \\
    Scheduler & Cosine & Cosine\\
    Epochs & 1 & 1 \\
    Batch Size & 4 & 4\\
    Hours & 20 & 10\\
    \bottomrule
  \end{tabular}
  \caption{Training settings.}
  \label{tab:Hyperparameter}
\end{table}

\begin{table}
  \centering
  \begin{tabular}{@{}l|cccccccccc}
    \toprule
    Dataset &  Type & Samples\\
    \midrule
    GeoChat \cite{kuckreja2024geochat} & IF & 216k\\
    VRSBench \cite{li2024vrsbench} & IF & 141k\\
    FIT-RS \cite{luo2024skysensegpt} & IF & 160k\\
    \midrule
    RSteller \cite{ge2024rsteller} & IC & 100k \\
    RSVQA-LR \cite{lobry2020rsvqa} & VQA & 57k\\
    RSVQA-HR \cite{lobry2020rsvqa} & VQA &100k\\
    RSITMD \cite{yuan2022exploring} & SC  & 2k\\
    RSICD \cite{lu2017exploring}  & SC  & 1k\\
    UCM \cite{yang2010bag} & SC  & 2k\\
    NWPU \cite{cheng2017remote} & SC  & 5k\\
    fMoW \cite{christie2018functional}  & SC  & 53k\\
    DIOR-RSVG \cite{zhan2023rsvg} & VG  & 30k\\
    OPT-RSVG \cite{li2024language} & VG  & 24k\\
    LEVIR-MCI \cite{liu2024change} & CC \& CD  & 40k\\
    Potsdam & SS  & 12k\\
    UAVid \cite{lyu2020uavid} & SS  & 8k\\
    LoveDA \cite{wang2021loveda} & SS  & 8k\\
    OpenEarthMap \cite{xia2023openearthmap} & SS & 14k\\
    landcover \cite{boguszewski2021landcover} & SS & 12k\\
    \midrule
    all &  - & 985k \\
    \bottomrule
  \end{tabular}
  \caption{Detailed statistics of multi-task RS datasets. Abbreviations IC, IF, VQA, SC, VG, SS, CC, CD, in the table stand for image captioning, instruction-following, visual question answering, scene classification, visual grounding, semantic segmentation, change captioning and change detection, respectively.}
  \label{tab:multi-taskRS}
\end{table}

\begin{table}
  \centering
  \begin{tabular}{@{}l|cccccccccc}
    \toprule
    Dataset &  Type & Samples\\
    \midrule
    ShareGPT4V \cite{chen2023sharegpt4v} & Captioning & 100k \\
    AllSeeingV2 \cite{wang2025all} & Grounding VQA & 200k \\
    DVQA \cite{kafle2018dvqa} & Diagram VQA & 40K \\
    ChartQA \cite{masry2022chartqa} & Diagram VQA & 3K \\
    AI2D \cite{hiippala2021ai2d} & Diagram VQA & 2K \\
    DocVQA \cite{mathew2021docvqa} & Document VQA & 2K \\
    COCO stuff \cite{caesar2018coco} & Semantic Segmentation & 200K \\
    \midrule
    all & - &  547k \\
    \bottomrule
  \end{tabular}
  \caption{Detailed statistics of general instruction dataset.}
  \label{tab:general instruction}
\end{table}

\begin{table}
  \centering
  \begin{tabular}{@{}l|cccccccccc}
    \toprule
    Granularity &  Dataset & Type\\
    \midrule
    \multirow{8}{*}{Image-level} & RSITMD & SC\\
     & RSICD \cite{yuan2022exploring}& SC\\
     & UCM \cite{yang2010bag}& SC\\
     & NWPU \cite{cheng2017remote}& SC\\
     & fMoW \cite{christie2018functional}& SC\\
     & RSVQA \cite{lobry2020rsvqa}& VQA\\
     & VRSBench-Cap \cite{li2024vrsbench}& CAP \\
     & LEVIR-MCI \cite{liu2024change} & CC \\
     \midrule
    \multirow{4}{*}{Region-level} & DIOR-RSVG \cite{zhan2023rsvg}& VG \\
     & OPT-RSVG \cite{li2024language}& VG \\
     & VRSBench-Ref \cite{li2024vrsbench}& VG \\
     & GeoChat$^{\ast}$ \cite{kuckreja2024geochat} & REG \\

    \midrule
    \multirow{6}{*}{Pixel-level} & Potsdam & SS \\
        & UAVid \cite{lyu2020uavid}& SS \\
        & LoveDA \cite{wang2021loveda}& SS \\
        & OpenEarthMap \cite{xia2023openearthmap}& SS \\
        & landcover \cite{boguszewski2021landcover}& SS \\
        & LEVIR-MCI \cite{liu2024change}& CD \\
    \bottomrule
  \end{tabular}
  \caption{Detailed statistics of datasets in second stage training.}
  \label{tab:stage2data}
\end{table}

\section{Hyperparameter}
\label{sec:Hyperparameter}
We trained all models on 4 Nvidia A40 GPUs (40GB). we employ the AdamW optimizer along with a learning rate of 0.00005, warmup ratio of 0.03 and weight decay of 0. Both two training stages last for 1 epoch, with a batch size of 4. See \cref{tab:Hyperparameter} for detailed training settings.

\section{Prompt for Evaluation}
In this section, we describe the prompts we use when evaluating each task in detail.

\paragraph{Scene Classification}
\begin{itemize}
\item \textless image\textgreater \textbackslash n[CLS] Choose the best category describe the image from: \{\emph{class names}\}. Only output the category.
\end{itemize}

\paragraph{Visual Question Answering}
\begin{itemize}
\item \textless image\textgreater \textbackslash n[VQA] \{\emph{question}\}.
\end{itemize}

\paragraph{Visual Grounding}
\begin{itemize}
\item \textless image\textgreater \textbackslash n[VG] \{\emph{object describing}\}.
\end{itemize}

\paragraph{Semantic Segmentation}
\begin{itemize}
\item \textless image\textgreater \textbackslash n[SEG] Can you segment the \{\emph{class name}\} in this image?
\item \textless image\textgreater \textbackslash n[SEG] Please segment the \{\emph{class name}\} in this image.
\item \textless image\textgreater \textbackslash n[SEG] What is \{\emph{class name}\} in this image? Please respond with segmentation mask.
\item \textless image\textgreater \textbackslash n[SEG] What is \{\emph{class name}\} in this image? Please output segmentation mask.
\end{itemize}

\paragraph{Change Captioning}
\begin{itemize}
\item \textless image\textgreater  \textless image\textgreater \textbackslash n[CCD] Please briefly describe the changes in these two images.
\item \textless image\textgreater  \textless image\textgreater \textbackslash n[CCD] What are the differences between these two images?
\item \textless image\textgreater  \textless image\textgreater \textbackslash n[CCD] Can you describe the key changes in the second image compared to the first?
\item \textless image\textgreater  \textless image\textgreater \textbackslash n[CCD] What visual changes can be observed between the two images?
\item \textless image\textgreater  \textless image\textgreater \textbackslash n[CCD] Highlight the main differences between the two images.
\item \textless image\textgreater  \textless image\textgreater \textbackslash n[CCD] What has been altered in the second image compared to the first one?
\item \textless image\textgreater  \textless image\textgreater \textbackslash n[CCD] Describe any noticeable transformations in these images.
\item \textless image\textgreater  \textless image\textgreater \textbackslash n[CCD] What specific details have changed from the first image to the second?
\item \textless image\textgreater  \textless image\textgreater \textbackslash n[CCD] Point out the adjustments made between these two images.
\item \textless image\textgreater  \textless image\textgreater \textbackslash n[CCD] What are the most significant changes visible in the images?
\end{itemize}

\paragraph{Change Detection}
\begin{itemize}
\item \textless image\textgreater \textless image\textgreater \textbackslash n[SEG] Please Segment the building area in the images that have changed in the second image..
\end{itemize}

\section{More Qualitative Results}
\label{sec:moreQualitative}
We present more qualitative results in \cref{fig:suppl_vg}, \cref{fig:suppl_seg}, \cref{fig:suppl_ccd}, which presents the visual grounding, semantic segmentation and change analysis performance. In addition, we have attached a demo video that contains richer examples.


\end{document}